\documentclass[journal]{IEEEtran} 

\usepackage[
    backend=biber,
    style=ieee,
    sorting=none,
    sortcites=true
]{biblatex}
\usepackage{graphicx} 
\usepackage{glossaries}
\usepackage{booktabs,longtable,tabularx,makecell,array} 
\usepackage{lipsum}
\usepackage{subcaption}
\usepackage{float}
\usepackage{microtype}
\usepackage{caption}
\title{Manual, Joystick, or Haptic Control? An In Vitro Comparison of Navigation Strategies for Robotic Interventional Neuroradiology Procedures}

\author{
        Benjamin~Jackson\textsuperscript{1},
        Nikola~Fischer\textsuperscript{1}, 
        Harry~Robertshaw\textsuperscript{1}, 
        Xingyu~Chen\textsuperscript{1}, 
        S.M.Hadi~Sadati\textsuperscript{1,2}, 
        Yang~Li\textsuperscript{1}, \\
        Jeremy~Lynch\textsuperscript{3}, 
        Nasr~Abdelsalam\textsuperscript{4}, 
        Jonathon~Buwanabala\textsuperscript{3}, 
        Matthew~Benger\textsuperscript{3}, 
        Sara~Sciacca\textsuperscript{3}, 
        Naga~Kandasamy\textsuperscript{3}, \\
        Marco~Mancuso-Marcello\textsuperscript{5}, 
        Parthiban~Balasundaram\textsuperscript{3}, 
        Sahan~Guruge\textsuperscript{3}, 
        Neelan~Das\textsuperscript{6}
        Alejandro~Granados\textsuperscript{1}, 
        Kawal~Rhode\textsuperscript{1}, 
        and~Thomas~C~Booth\textsuperscript{1,}\textsuperscript{3}

    \IEEEauthorblockA{\textsuperscript{1}School of Biomedical Engineering \& Imaging Sciences, King's College London, UK}
    
    \IEEEauthorblockA{\textsuperscript{2}School of Engineering and Materials Science, Queen Mary University of London, UK}
    
    \IEEEauthorblockA{\textsuperscript{3}Department of Neuroradiology, King's College Hospital NHS Foundation Trust, London, UK}
    
    \IEEEauthorblockA{\textsuperscript{4}The Walton Centre NHS Foundation Trust, Liverpool, UK}

    \IEEEauthorblockA{\textsuperscript{5}University College London Hospitals NHS Foundation Trust, London, UK}

    \IEEEauthorblockA{\textsuperscript{6}East Kent Hospitals University NHS Foundation Trust, Kent, UK}

    \IEEEauthorblockA{Corresponding author: (thomas.booth@kcl.ac.uk)}
}

\newacronym{mt}{MT}{mechanical thrombectomy}
\newacronym{threed}{3D}{three-dimensional}
\newacronym{DoF}{DoF}{degree of freedom}
\newacronym{Hz}{Hz}{hertz}
\newacronym{sla}{SLA}{stereolithography}
\newacronym{cad}{CAD}{computer-aided design}
\newacronym{cam}{CAM}{computer-aided manufacturing}
\newacronym{aic}{AIC}{Akaike Information Criterion}
\newacronym{cv}{CV}{coefficient of variation}
\newacronym{stl}{STL}{stereolithography}
\newacronym{kch}{KCH}{King’s College Hospital}
\newacronym{sie}{SIE}{Surgical \& Interventional Engineering}
\newacronym{ica}{ICA}{internal carotid artery}
\newacronym{bca}{BCA}{brachiocephalic artery}
\newacronym{cca}{CCA}{common carotid artery}
\newacronym{rc}{RC}{robotic controller}

\begin{document}


\maketitle

\begin{abstract}
Objective: To evaluate robotic controller interfaces for interventional neuroradiology procedures in-vitro incorporating a force-sensing platform to assess safety.

Methods: A custom endovascular robot, device-mimicking controller, and sensorized neurovascular phantom were developed. Ten interventional neuroradiologists (4 novices, 6 experts) performed simulated navigations using four control modalities: device-mimicking controllers with and without haptic feedback, joystick-based input, and manual navigation. Navigation time, peak vessel-wall forces, incorrect catheterisations, and prolapse events were assessed, alongside user analyses.

Results: Manual navigation was fastest (mean 47.7 s) compared to haptic-on (248.7 s), haptic-off (314.7 s), and joystick (392.6 s) modalities (p$<$0.001). Regardless of controller type, vessel-wall forces were below the 0.70 N puncture threshold; therefore all modalities were considered safe. Joystick produced significantly more prolapse events than manual control (1.56 vs 0.13; p=0.018). Operator experience was relevant to performance: experts made fewer incorrect catheterisations than novices (0.25 vs 0.62; p=0.035) and applied less vessel-wall force (p$<$0.0005); these effects were sustained across controllers but accentuated when haptics were on. Users perceived haptic on and haptic off as similarly intuitive, and more intuitive than joystick (p=0.033).

Conclusion: Device-mimicking robotic controllers outperform joystick interfaces on most metrics; haptic feedback shows promising but non-significant performance benefits.
\end{abstract}

\begin{IEEEkeywords}
Endovascular robotics, interventional neuroradiology, haptic feedback, vascular phantom, force sensing, teleoperation, mechanical thrombectomy.
\end{IEEEkeywords}

\section{Introduction}

Stroke remains a significant public health burden, accounting for approximately 6.55 million deaths in 2019~\cite{Feigin2021}. While mechanical \gls{mt} is the standard treatment for large-vessel occlusion, its reach is restricted by procedural complexity and a shortage of trained interventional neuroradiologists~\cite{BENDSZUS20231753}. Robotic systems offer a potential solution by increasing precision and enabling tele-operated remote intervention, though a translational gap remains for operators between manual manipulation of devices in the operating room (OR), and using robotic interfaces to perform that task~\cite{Crinnion539}. Robotic interfaces generally utilize either non-isomorphic interfaces (e.g., joysticks) or device-mimicking interfaces (i.e. replicating the interactions clinicians have with common endovascular devices). Device-mimicking has the potential to preserve operator intuition and reduce the translational gap~\cite{Jackson2023}.

Comparative verification of robotic interface control requires assessment of metrics such as navigation time, path efficiency, and force exerted from the endovascular tools on the vascular anatomy. While these have been evaluated \textit{in silico}~\cite{Jackson2023}, physical validation in high-fidelity substrates is necessary to accurately reproduce the tactile conditions encountered in clinical settings when making these comparative assessments.

Relatedly, reproducing the tactile conditions through haptics in \gls{mt} robotics is a developing field~\cite{Jackson2023}, following a broader shift toward force-reflective interfaces in robotic surgery, such as those integrated into recent Da Vinci systems~\cite{davinci2025}.

Another consideration when comparing robotic interfaces and endovascular robot safety relates to vessel wall damage during actuation. When using off-the-shelf wires and catheters, safety can be assessed \textit{in silico} by measuring forces at the device tip as a surrogate for tissue damage~\cite{Jackson2023, Robertshaw2025}. While these simulations use force estimates as a metric for optimization, quantifying these interactions \textit{in vitro} is challenging. While traditional physical phantoms often lack the ability to provide objective force sensing, Fischer \textit{et al.}~\cite{Fischer2023} addressed this by developing modular, \gls{threed}-printed vascular phantoms with integrated piezoresistive sensors. These platforms allow for the measurement of instrument--wall contact at clinically relevant sites.

The aim of this work is to perform a comparative verification of control methodology for robotic interventional neuroradiology procedures \textit{in vitro}. To achieve this, our contributions are: (1) we developed a sensorized phantom that allows for device tip force assessment during robotic navigation; (2) we compared the performance of robotic controllers, including those with haptic feedback, against a manual control reference; and (3) we assessed the impact of interventional neuroradiology operator experience by comparing novice and expert performance across these control modalities.

\section{Methods}
\label{sec:methods}

A robotic platform was developed to allow comparative verification of control methodology for robotic interventional neuroradiology procedures \textit{in vitro}. The system consisted of three core components: a purpose-built device-mimicking controller (that replicated the interactions clinicians have with common endovascular devices) which could be used with or without haptics; a custom-built endovascular robot; and a multi-node force-sensing neurovascular phantom.

\subsection{Robotic Controller with Haptic Feedback}
\label{sec:methods:rc}

\begin{figure}[ht]
    \centering
    \includegraphics[width=0.49\textwidth]{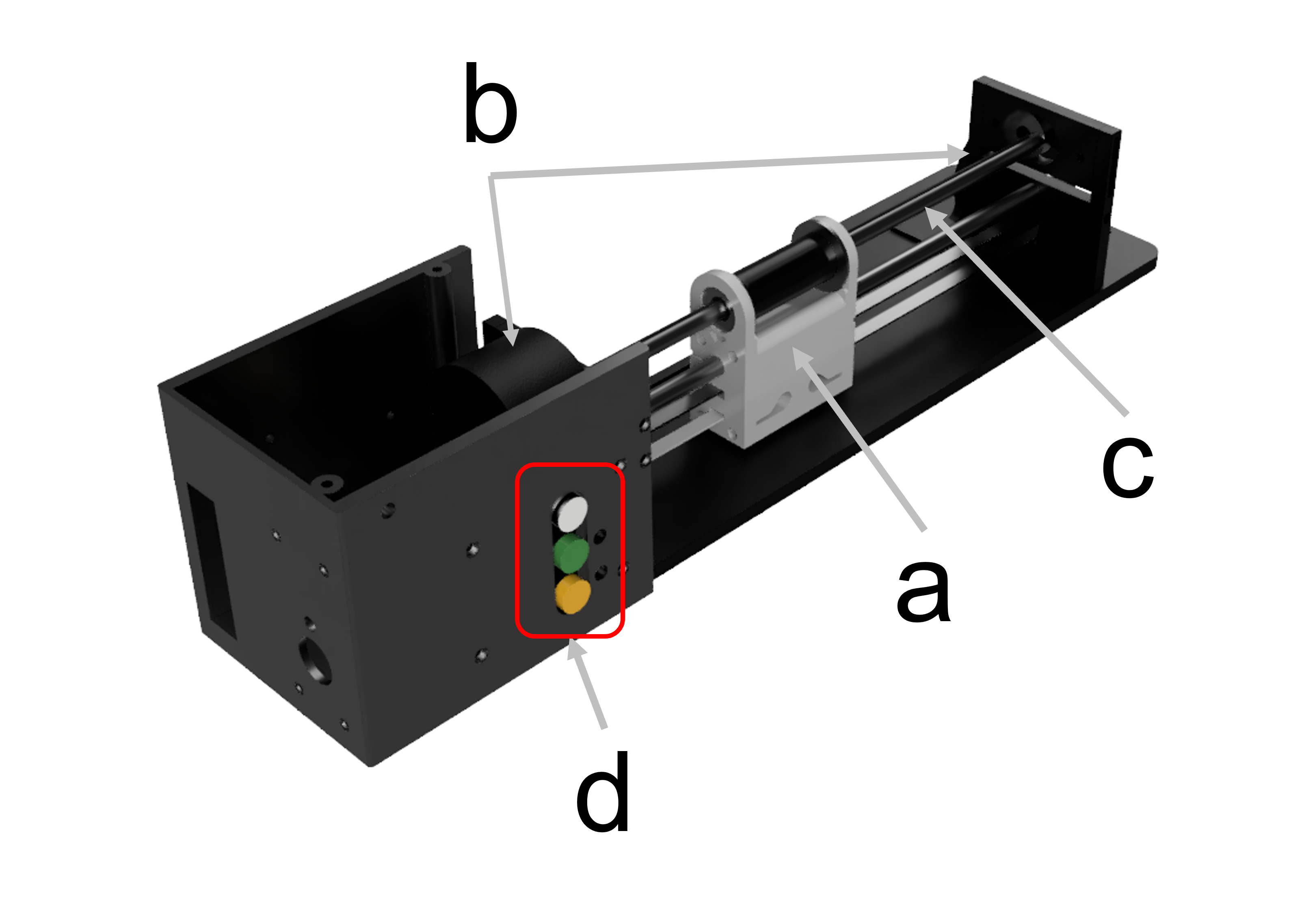}
    \caption{Components of our dual-device robotic controller for interventional neuroradiology procedures: 
    (\textbf{a}) Device-mimicking user handle. 
    (\textbf{b}) Brushless actuation motors. 
    (\textbf{c}) Force-feedback transmission pathway for haptic rendering.
    (\textbf{d}) Mode of operation - \textbf{(White button)} Control both the catheter and guidewire simultaneously, \textbf{(Green button)} Control only the guidewire, \textbf{(Yellow button)} Control only the catheter.
    }
    \label{fig:haptic_controller}
\end{figure}

The robotic controller was adapted from a previous study~\cite{Jackson2023} by increasing the haptic feedback driver frequency to 250 Hz and upgrading the haptic motors to MyActuator RMD-L 4015 (Suzhou Micro Actuator Technology Co., Huaqiao, China). The system supported both translation and rotation of endovascular devices and, when enabled, delivered force feedback through a force-reflective interface. Enabling and disabling haptic feedback in different experiments allowed performance related to the different modes to be compared. 

Haptic feedback was implemented in a closed-loop scheme: axial and torsional forces acting on the guidewire were measured by distal sensors, processed, and transmitted to the controller. These signals were then mapped to the corresponding user inputs (translation and rotation), with motor currents modulated to render proportional resistive forces at the handle. In this way, operators experienced an approximation of in-vessel loading as torques and forces opposing their input, enabling direct haptic correspondence between device–phantom interaction and manual control.

\paragraph*{Controller Architecture}  
The device-mimicking controller incorporated brushless motors and high-frequency control electronics to enable stable and realistic haptic rendering at 250 Hz. Fig.~\ref{fig:haptic_controller} illustrates the main components.

\subsection{Endovascular Robot}
\label{sec:methods:robot}

\begin{figure}[ht]
    
    \centering
    \includegraphics[width=0.49\textwidth]{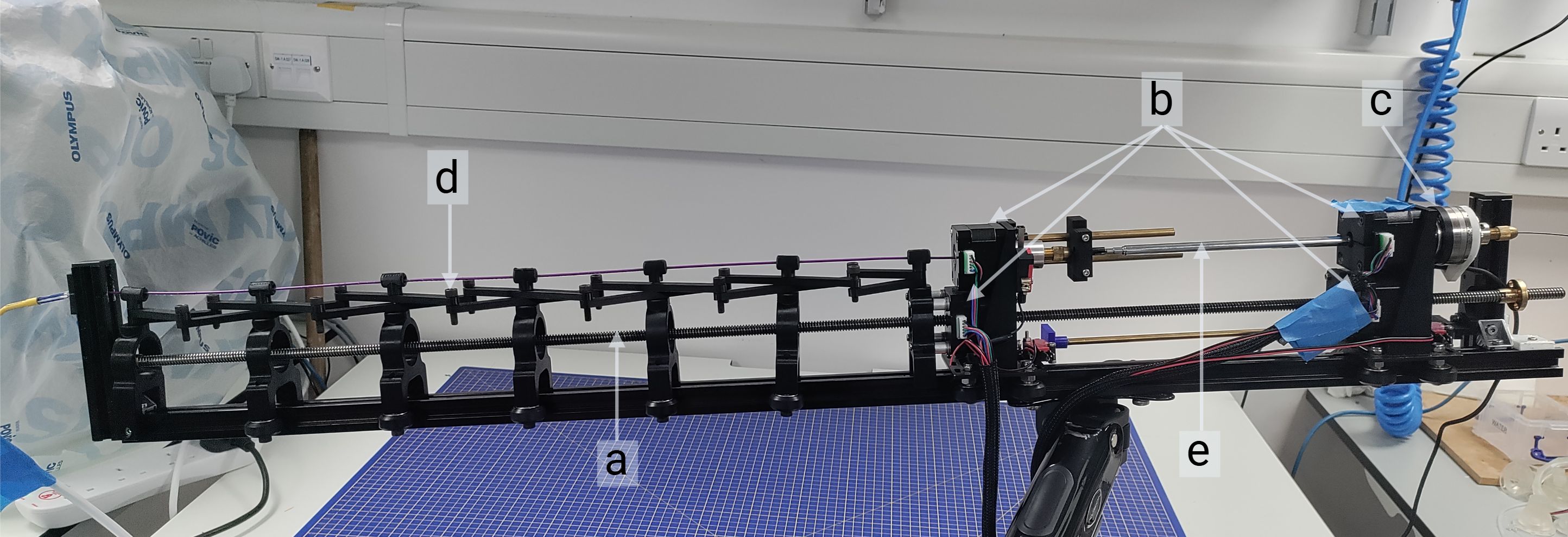}
    \caption{Schematic of the endovascular robot. 
    (a) Lead screw mechanism for translation. 
    (b) Nema 17 stepper motors for translation and rotation. 
    (c) Guidewire manipulation carriage with integrated ATI 6-\gls{DoF} force sensor. 
    (d) Scissor-frame anti-buckling system between the front and the first carriage. 
    (e) Telescopic tube anti-buckling system between the first and second carriage.}
    \label{fig:endovascular_robot}
\end{figure}

The endovascular robot was developed to provided precise control of concentric neuroendovascular devices and to support the transition from purely \emph{in silico} to physical phantom experimentation. The system delivered two \gls{DoF}: translation and rotation. A precision lead screw mechanism enabled high-resolution linear advancement, while concentric shafts allow simultaneous manipulation of up to three devices, including guidewires, micro-catheters, and stent retrievers \cite{sedati2025, KCLRoboticEndovascular2026}.

\paragraph{Design and Actuation}  
Device motion was powered by two Nema 17 stepper motors (24 V, 600 mA, 80 Ncm): one dedicated to translation along the lead screw and the other to rotational control. This dual-motor arrangement ensures smooth, independent movements even within tortuous neurovascular geometries.

\paragraph{Force Measurement}  
To capture interaction forces for haptic rendering, an ATI Mini 58 six-\glsfirst{DoF} force sensor (ATI, Ontario, Canada) was integrated into the guidewire manipulation carriage (Fig.~\ref{fig:endovascular_robot}). This placement allowed direct measurement of axial, lateral, and torsional forces applied to the proximal end of the guidewire. These data were sampled at 250 Hz and provide the input necessary for real-time haptic feedback in the controller.

\paragraph{Anti-Buckling Systems}  
Two mechanical support systems (scissor-frame and telescopic tube, Fig.~\ref{fig:endovascular_robot}(d-e)) constrained endovascular devices axially and laterally, preventing buckling while permitting smooth translation and rotation.

\subsection{Granular Force Sensing Phantom}

\begin{figure}[ht]
    \centering
    \includegraphics[width=0.49\textwidth]{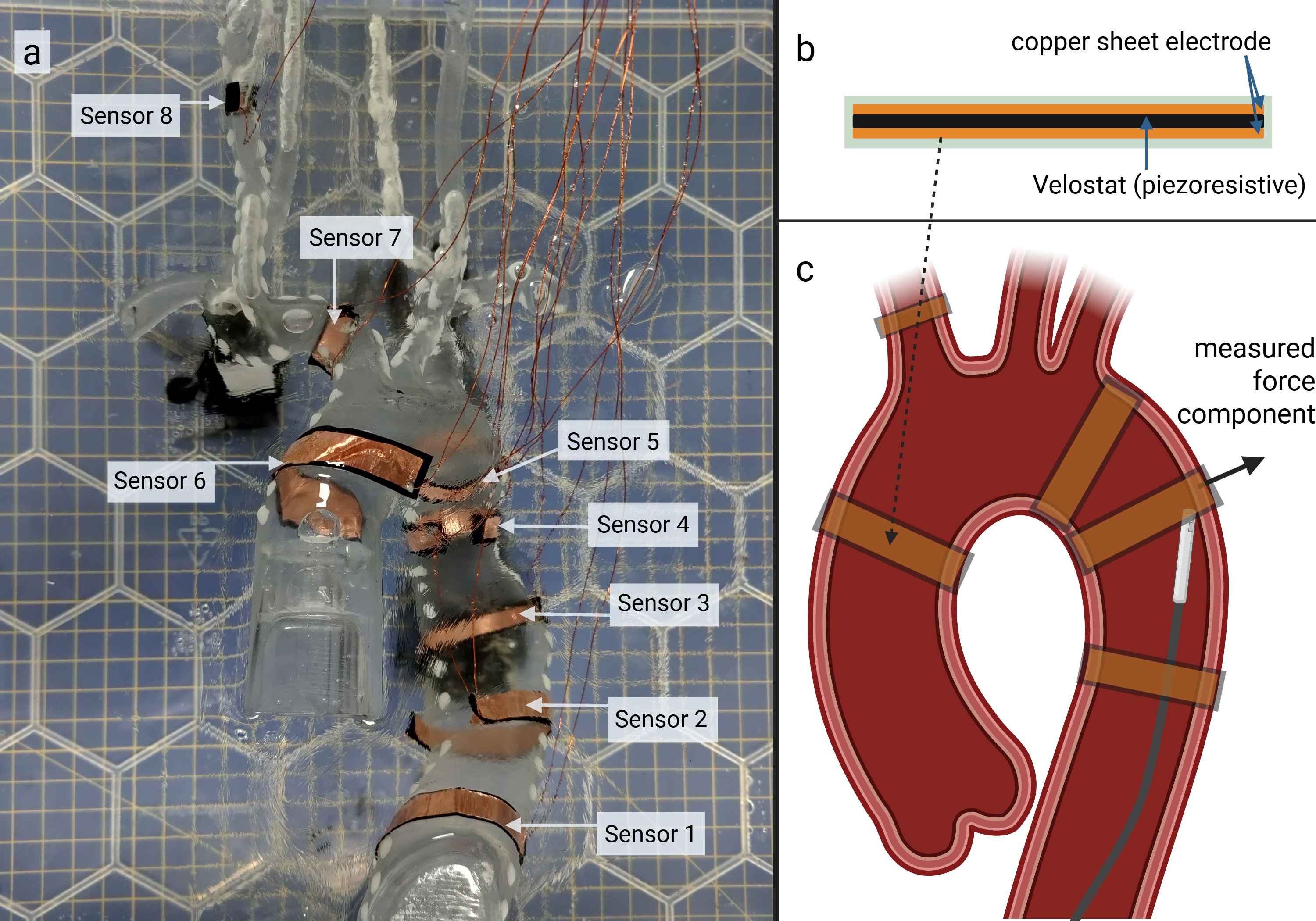}
    \caption{(a) sensorized endovascular phantom with the position of the 8 force sensors (b) piezoresistive force sensors (c) interaction of endovascular device with sensorized phantom.}
    \label{fig:force_sensing_phantom}
\end{figure}

\paragraph{Phantom Design Objectives}
The phantom was designed to (i) reproduce patient-specific vascular geometry at interventional resolution, (ii) permit direct visualization of device--wall interactions, (iii) quantify local interaction forces of the device and vessel wall at clinically relevant sites, and (iv) support repeatable, controlled experimentation with the endovascular robot (Section~\ref{sec:methods:robot}). 

\paragraph{Phantom Imaging-to-Print Pipeline}
The vascular geometry was extracted from anonymized patient imaging data collected at \gls{kch} NHS Foundation Trust (UK Research Ethics Committee 24/LO/0057). Segmentation of the relevant vessels was performed using \emph{3D Slicer}~\cite{Fedorov2012}. The resulting surface model was exported in \gls{stl} format. Vessel walls were uniformly thickened to [\textit{t} = 1.4\,mm] to enable 3D printing and repeated handling without compromising luminal fidelity. This workflow preserved the patient-specific morphology while ensuring structural robustness for experimental navigations.

\paragraph{Phantom Fabrication and Post-Processing}
The phantom (Fig.~\ref{fig:force_sensing_phantom}) was fabricated from flexible clear photopolymer resin (Siraya Tech Tenacious, Siraya Tech, Tainan, Taiwan) using a Saturn 5 Ultra \gls{sla} printer (Elegoo, Shenzhen, China). For experimental use, the phantom was submerged in glycerin (Hexeal, Norwich, United Kingdom), which served as a static blood analog in place of a flow model. In clinical procedures, contrast agent is injected into flowing blood to highlight the vascular lumen by applying a projected mask on the screen (typically called a roadmap). To enable operators to visualize the vascular lumen, small deposits of radiopaque barium sulfate were applied externally (Fig.~\ref{fig:experiment_setup}(c)). Although this produced an image appearance different from clinical angiography, it conveyed a comparable amount of navigational information - i.e. vessel outline - while maintaining experimental repeatability. 

\paragraph{Granular Force Sensing Architecture}
\label{sec:force_sensing_phantom}
Wall contact forces were measured using custom-built piezoresistive sensors based on Velostat film, a conductive polymer material whose resistance decreases under pressure \cite{polym12122905}. Each sensing element consisted of a Velostat sheet sandwiched between two copper tape electrodes acting as anode and cathode (Fig.~\ref{fig:force_sensing_phantom}b), following the approach described by~\cite{Fischer2021}. To enable submersion in glycerin, which was required for use as a static blood analog, the sensors were encapsulated in a thin layer of flexible silicone (Smooth-On Dragonskin 10, Pennsylvania, USA), ensuring both electrical insulation and mechanical conformity to the vessel wall. Similar Velostat-based force sensors have recently been applied in other vascular phantom platforms~\cite{Fischer2023}. The encapsulated design provided robustness under repeated loading, allowed consistent force readout in submerged conditions, and enabled the generation of a spatial force map during device navigation.

\paragraph{Vascular Sensor Placement}
Sensor sites were identified from prior \emph{in silico} force distributions \cite{Jackson2023}, selecting anatomically feasible locations with the highest cumulative forces along the vessel centerline while respecting manufacturing constraints.

\paragraph{Sensor Electronics and Sampling}
Eight encapsulated sensors were interfaced via Arduino Uno microcontroller with signal conditioning and analog-to-digital conversion. Data were acquired at 100\,Hz with retrospective digital low-pass filtering to suppress noise while preserving force signal fidelity.

\paragraph{Calibration and Metrology}
Sensors were calibrated using quasi-static loads (0–200 mN) with fits selected via AIC minimization. Median calibration $R^2$ was 0.62; repeatability at 100 mN was 14.8\% CV; nearest-neighbor cross-talk was <12\%. Per-site force uncertainty was 28 mN at 95\% confidence.

\medskip
In summary, the clear-resin, sensorized neurovascular phantom provided a controllable, instrumented substrate for evaluating robotic navigation with different controller interfaces. Coupled with the high-rate robot forcesensing (Section~\ref{sec:methods:robot}), it enabled site-specific, quantitative assessment of device--vessel interactions under repeatable stationary fluidic conditions, supporting the safety and performance analyses presented in Section~\ref{sec:experiments}.

\section{Experiments}
\label{sec:experiments}

We compared the performance of different robotic system controllers with conventional manual endovascular navigation with particular emphasis on assessing the contribution and perceived validity of haptic feedback.

\subsection{Participants and Setting}
\begin{figure}[ht]
    \centering
    \includegraphics[width=0.49\textwidth]{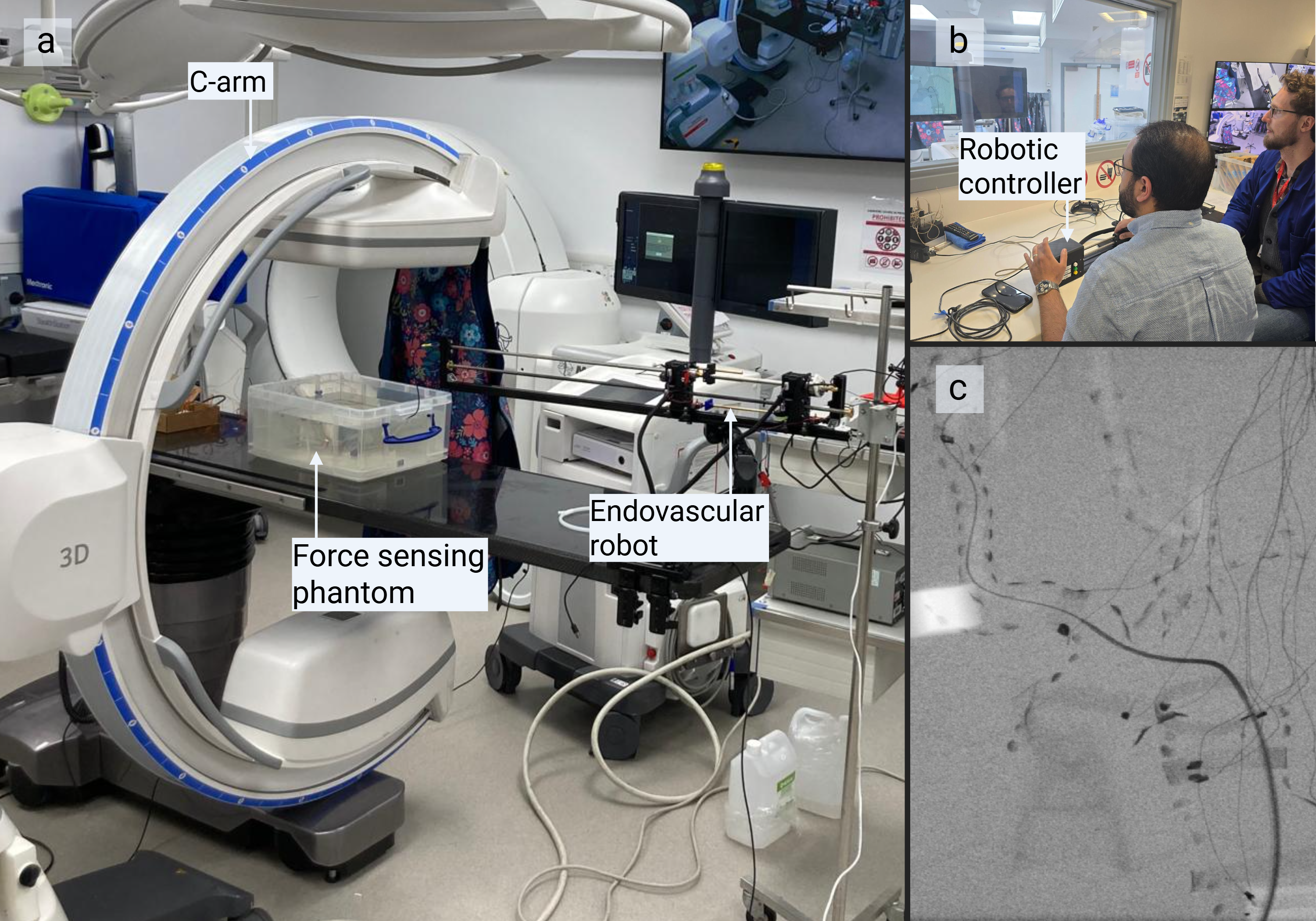}
    \caption{Experimental setup.  
    (a) The robotic system and vascular phantom positioned in a mock angiography suite with a Siemens mobile C-arm (Siemens Healthineers, Erlangen, Germany).  
    (b) Control room showing a participant operating the teleoperated robot via the custom interface.  
    (c) Example fluoroscopic image acquired during navigation. Barium deposits are added to the phantom to enhance visualization of the vascular lumen.}
    \label{fig:experiment_setup}
\end{figure}

Experiments were conducted in the mock operating room at the \gls{sie} Laboratory, King’s College London, UK. 
Ten interventional neuroradiologists participated, representing a range of experience: four novices (we defined as fewer than five years of independent practice) and six experts (more than five years of practice). 
Each participant completed a series of simulated neurovascular navigations using both robotic and manual approaches, within the experimental setup shown in Fig. ~\ref{fig:experiment_setup}.

\subsection{Experimental Conditions}
Each participant performed a total of eight navigations using a Terumo Glidewire (Terumo, Tokyo, Japan), Penumbra Slip-Cath Beacon (Penumbra, California, United States) and a fixed Penumbra Neuron MAX (Penumbra, California, United States) for stability.
These consisted of one navigation with each of the four controller configurations, followed by a repeat of all four to determine learning and consistency effects:
\begin{enumerate}
    \item \textbf{Manual:} standard manual navigation of devices without a robot.
    \item \textbf{Robotic controller (haptics on):} device-mimicking robotic controller with haptic feedback enabled.
    \item \textbf{Robotic controller (haptics off):} device-mimicking robotic controller without haptic feedback.
    \item \textbf{Joystick (Xbox, Microsoft, Washington, United States):} conventional joystick-based input serving as a non-device-mimicking baseline, in which button presses actuated the robot in a stepwise fashion.
\end{enumerate}

Before commencing the experimental navigation, participants were given five minutes of practice with each of the four navigation styles to familiarize themselves with the interface. 
During each trial, they were allowed a maximum of eight minutes to complete the navigation. 
If the target vessel was not reached within this limit, the navigation was terminated and a truncated completion time of eight minutes was recorded.  

\begin{figure}[ht]
    \centering
    \includegraphics[width=0.49\textwidth]{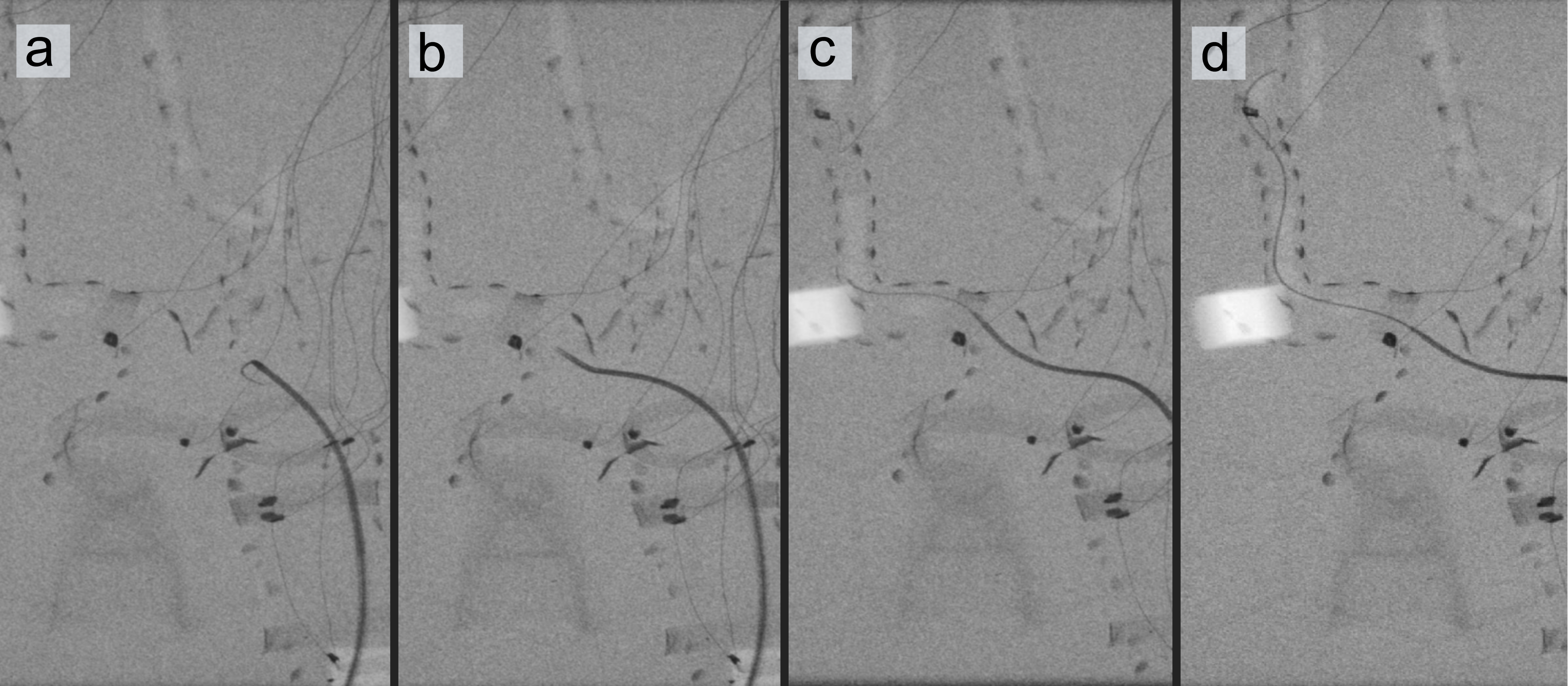}
    \caption{Fluoroscopic sequence of catheter navigation.  
    (a) Starting position at the aortic arch.  
    (b) Selective engagement of the \glsentrylong{bca}.  
    (c) Advancement into the \glsentrylong{cca}.  
    (d) Final position within the \glsentrylong{ica}.}
    \label{fig:fluoro_sequence}
\end{figure}

All participants were assigned the same intended navigation route (Fig.~\ref{fig:fluoro_sequence}). 
In each navigation, the guidewire and catheter were advanced through the descending aorta, into the \gls{bca}, then into the right \gls{cca}, and finally into the right \gls{ica}, as illustrated in Fig.~\ref{fig:fluoro_sequence}. 
The target endpoint was always the right \gls{ica}; participants were not required to advance further into the intracranial vasculature.

\subsection{Outcome Measures}
Outcomes were recorded to evaluate and compare navigation performance using the MUTUAL framework~\cite{Quarez2025-tl}:
\begin{itemize}
    \item \textbf{Navigation time:} total time in seconds to complete the planned route, further divided into subsegments corresponding to progression into the \gls{bca}, \gls{cca}, and \gls{ica}.
    \item \textbf{Vessel wall forces:} maximum forces applied to vessel walls. For each navigation, the peak force recorded by each sensor was extracted. A conservative safety threshold of 0.70\,N was adopted, based on reported puncture forces of approximately 0.72\,N for a 27-gauge stiff needle penetrating vascular tissue \cite{Siperstein2023}. Given that the guidewire and catheter are substantially more flexible than a needle, this threshold represents a highly conservative limit.
    \item \textbf{Incorrect catheterisations:} defined as catheter or guidewire insertion $>$10\,mm into a vessel not part of the planned route.
    \item \textbf{Prolapse events:} defined as a catheter or guidewire prolapsing out of the vessel lumen origin by $>$10\,mm.
\end{itemize}

\subsection{Qualitative Operator Feedback}
Additionally, qualitative operator feedback was gathered through a post-experiment questionnaire using a 10-point Likert scale. Questions evaluated the realism of the simulation, its usefulness for training, and the intuitiveness of each control method. The NASA-TLX~\cite{NASATLX} was asked for each control method, then participants were asked to choose their preferred control methodology and answer a 10-point System Usability Scale~\cite{inbook} assessing the system with their chosen controller.

\subsection{Statistics}
Data were analyzed using a one-way ANOVA to evaluate differences across control modalities, followed by Bonferroni-corrected post-hoc tests ($p_{adj}$) for multiple comparisons between sensor sites. To account for repeated measures and unequal variances between operator groups, Welch’s t-tests were performed to compare novice versus expert performance. We report both within-group and across-group variability via mean and standard deviation. For all analyses, statistical significance was defined as $p < 0.05$.

\section{Results}
\label{sec:results}

\subsection{Comparative Verification of Control Methods}
\label{subsec:control_verification}

Overall performance across all controllers (manual, haptic on, haptic off, joystick) was evaluated using four key metrics: navigation time, vessel-wall forces, incorrect catheterizations, and prolapse events.

\subsubsection{Time}
\label{subsubsec:cv_time}

 \begin{figure}[ht]
   \centering
   \includegraphics[width=0.47\textwidth]{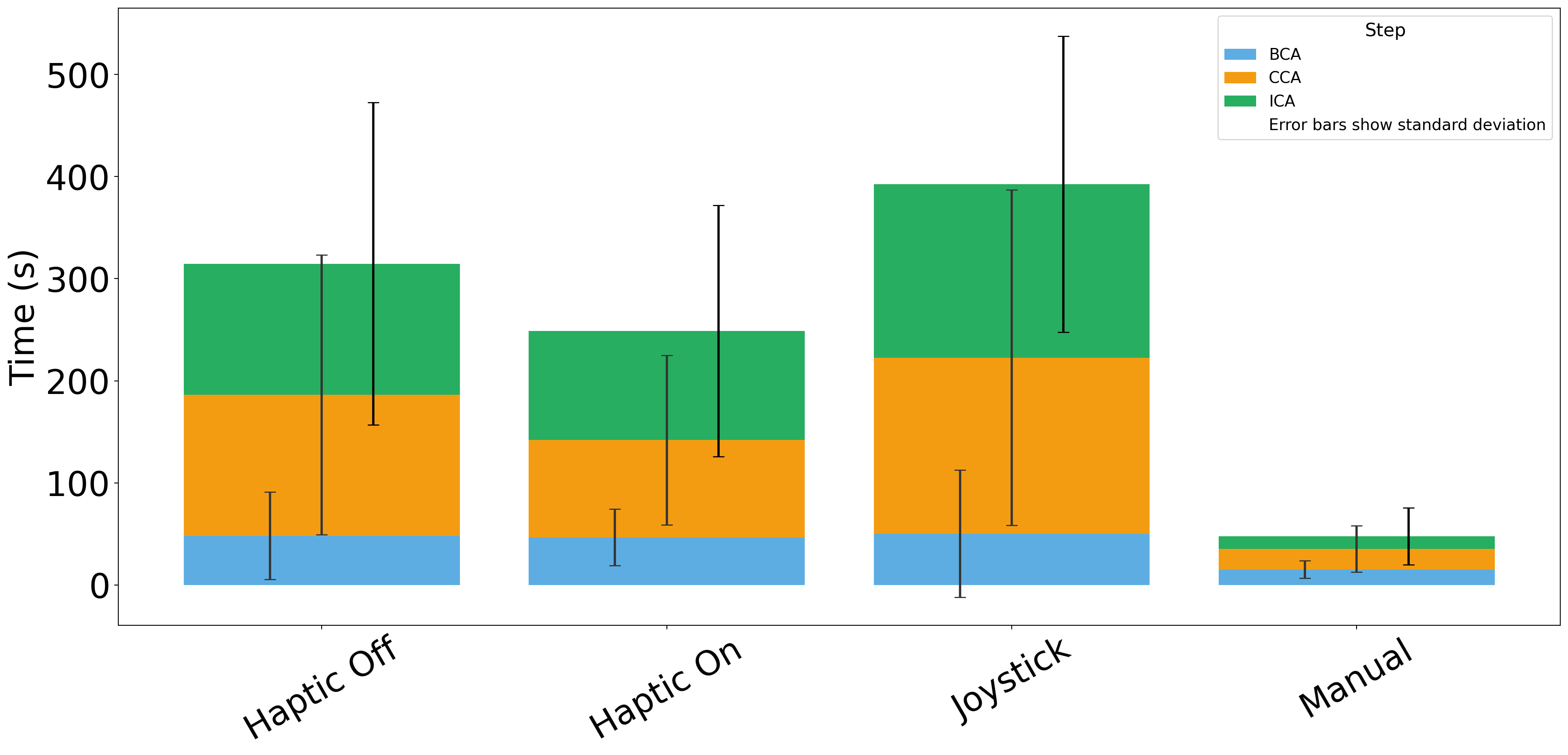 }
   \caption{Navigation times split into target checkpoints for the device-mimicking controller with the haptics on, the device-mimicking controller with haptics off, the joystick controller, and the manual navigation.}
   \label{fig:cv_time_by_controller}
 \end{figure}

Navigation times varied substantially across control methods (Fig.~\ref{fig:cv_time_by_controller}). 
Manual navigation was consistently the fastest, with a mean completion time of 47.7\,s. 
In contrast, all robotic controllers required considerably longer times: haptic on averaged 248.7\,s, haptic off 314.7\,s, and joystick navigation 392.6\,s. 
Among the robotic controllers, the haptic on mode yielded the shortest times, while joystick navigation was the slowest overall.  

The stacked bar plot in Fig.~\ref{fig:cv_time_by_controller} illustrates how times accumulated across anatomical segments (BCA, CCA, ICA). 
Across all controllers, the CCA and ICA contributed the largest share of navigation time, with joystick navigation showing particularly high variability.  

Statistical analysis confirmed a strong effect of controller type (one-way ANOVA, $F(3,65) = 24.77$, $p = 8.36 \times 10^{-11}$). 
Post-hoc comparisons indicated that manual navigation was significantly faster than each robotic controller ($p < 0.001$). 
Haptic feedback appeared the quickest robotic controller but differences between haptic on and haptic off did not reach statistical significance ($p = 0.47$).  

\subsubsection{Forces}

\begin{figure}[ht]
  \centering
  \includegraphics[width=0.47\textwidth]{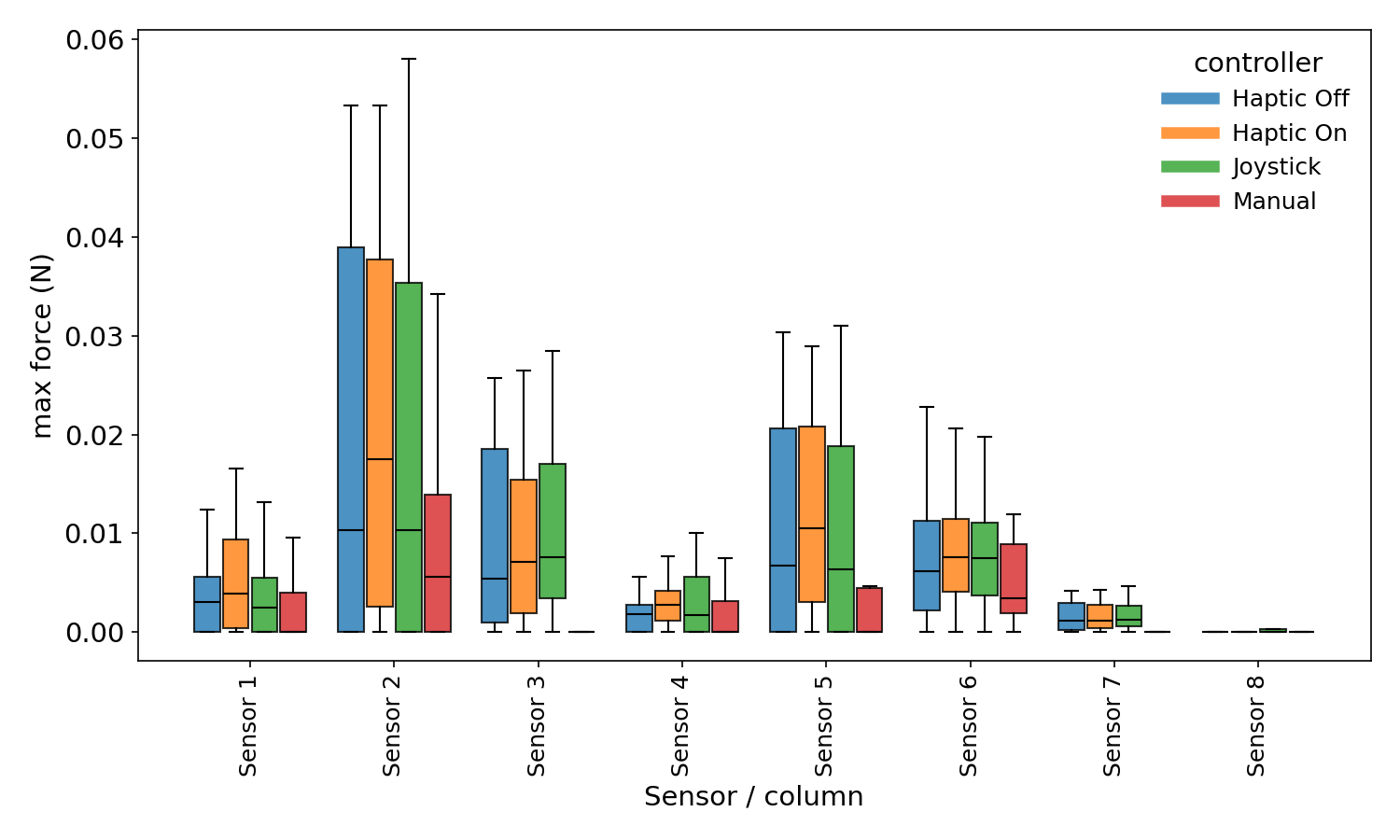}
  \caption{Per-sensor maximum force grouped by controller. Sensor placement found in Fig.~\ref{fig:force_sensing_phantom} Each sensor is plotted on the horizontal axis, with a separate box plot for each control method.}
  \label{fig:force_max_per_sensor}
\end{figure}

Maximum vessel-wall forces were analyzed across the eight on-phantom sensors and the four control methods. Fig.~\ref{fig:force_max_per_sensor} shows box plots of the per-sensor maximum force grouped by controller. Manual navigation consistently produced the lowest maximum forces across all sensors, with a mean value of $0.0032 \pm 0.0036$\,N. Whilst the robotic controllers (haptic off, haptic on and joystick) appeared to produce slightly higher forces—mean of $0.007$\,N - these were not significantly higher ($p = 0.33$). The highest forces were observed at sensors 2 and 3 (Fig.~\ref{fig:force_sensing_phantom}) for the robotic controllers, yet even these peak values were two orders of magnitude below the estimated puncture threshold of $0.70 \pm 0.29$\,N\cite{Siperstein2023}.

\subsubsection{Incorrect Catheterizations}
\label{subsubsec:cv_catheterisations}

\begin{figure}[ht]
    \centering
    \begin{subfigure}[t]{0.48\columnwidth}
        \centering
        \includegraphics[width=\linewidth]{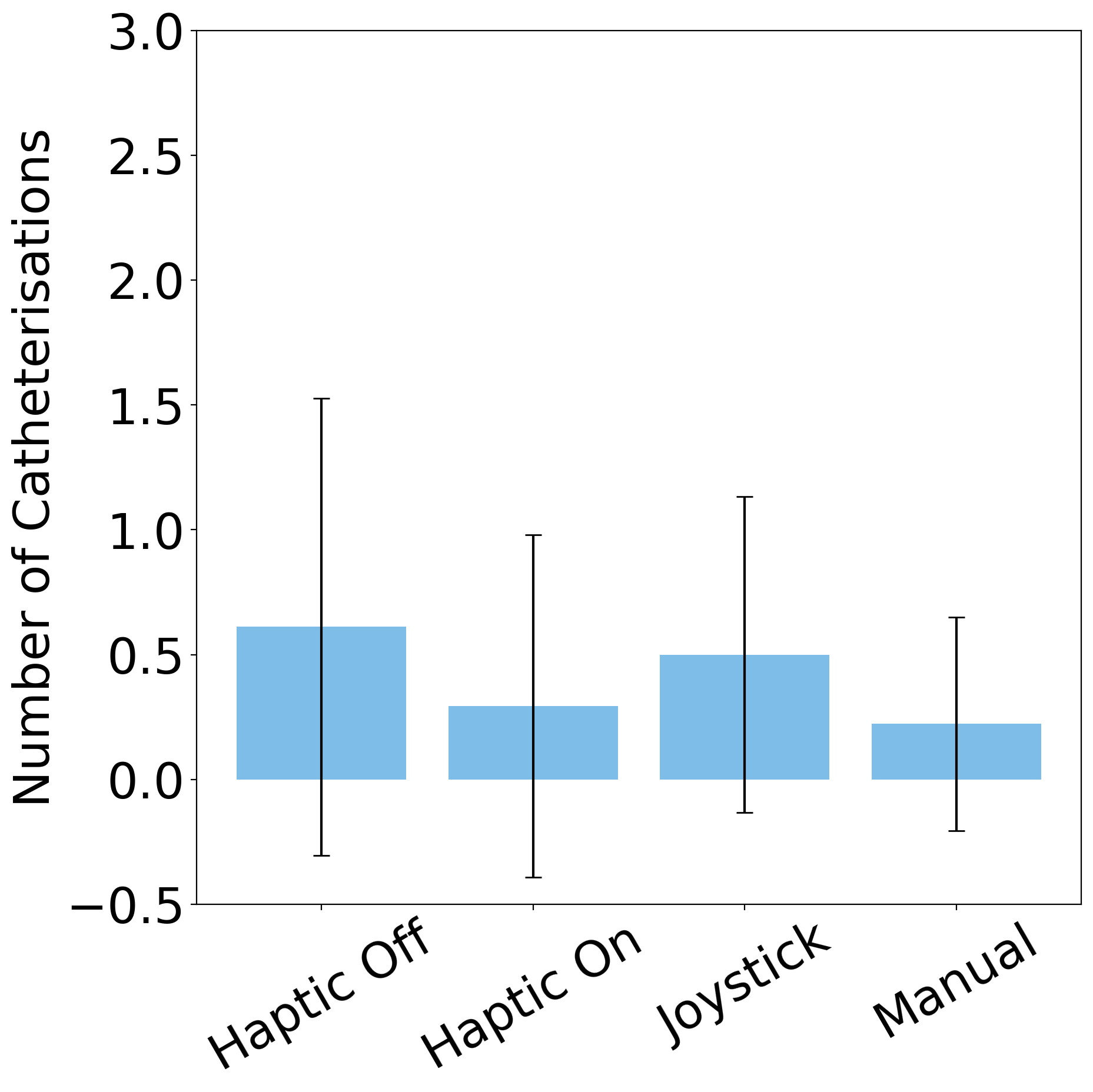}
        \caption{Mean number of incorrect catheterizations with error bars indicating standard deviation.}
        \label{fig:cv_catheterisations_by_controller}
    \end{subfigure}\hfill
    \begin{subfigure}[t]{0.48\columnwidth}
        \centering
        \includegraphics[width=\linewidth]{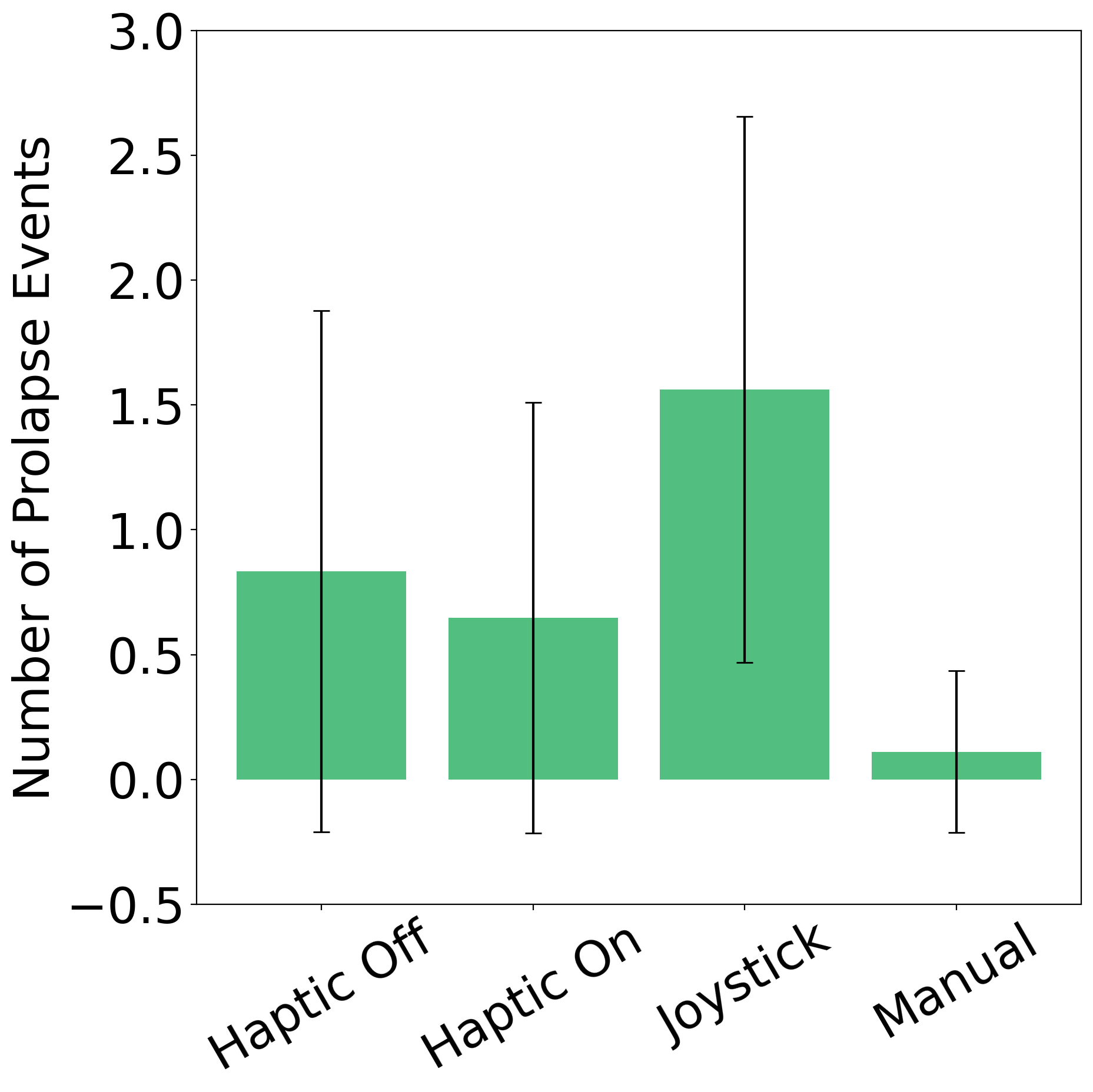}
        \caption{Mean number of prolapse events with error bars indicating standard deviation.}
        \label{fig:cv_prolapse_by_controller}
    \end{subfigure}

    \caption{Comparison of error-related outcomes during navigation (from descending aorta to \gls{ica}) across all control methods.}
    \label{fig:cv_error_outcomes}
\end{figure}

The frequency of incorrect catheterizations was generally low across all controllers (Fig.~\ref{fig:cv_catheterisations_by_controller}). 
Manual navigation produced the lowest error rate, with a mean of 0.22 incorrect catheterizations per navigation (from descending aorta to \gls{ica}). 
Among robotic modes, haptic on showed the fewest errors (0.29), followed by joystick navigation (0.50), while haptic off yielded the highest mean rate at 0.61.  

Despite these descriptive differences, statistical analysis did not reveal a significant effect of controller type (one-way ANOVA, $F(3,65) = 1.20$, $p = 0.315$).

\subsubsection{Prolapse Events}  
\label{subsubsec:cv_prolapse}  

The frequency of prolapse events varied between controller modes (Fig.~\ref{fig:cv_prolapse_by_controller}).  
Manual navigation had the lowest mean frequency, at only 0.13 ± 0.32 events per navigation (from descending aorta to \gls{ica}), while joystick navigation produced the highest frequency, with 1.56 ± 1.10 events per navigation.  
The haptic-assisted controllers fell in between: \textit{haptic off} yielded 0.83 ± 1.05 events, and \textit{haptic on} showed a slightly reduced mean of 0.64 ± 0.85 events.  

Statistical analysis confirmed significant differences between controllers (ANOVA, $p = 0.018$).  
Post-hoc tests indicated that joystick navigation was significantly more prone to prolapse compared with manual control ($p = 0.014$).  
Differences between the haptic controllers (on vs. off) were not significant, though a trend favored haptic feedback ($p = 0.26$).  


\subsection{Experts vs. Novices}
\label{subsec:nov_exp}

\begin{figure}[ht]
  \centering
  \includegraphics[width=0.47\textwidth]{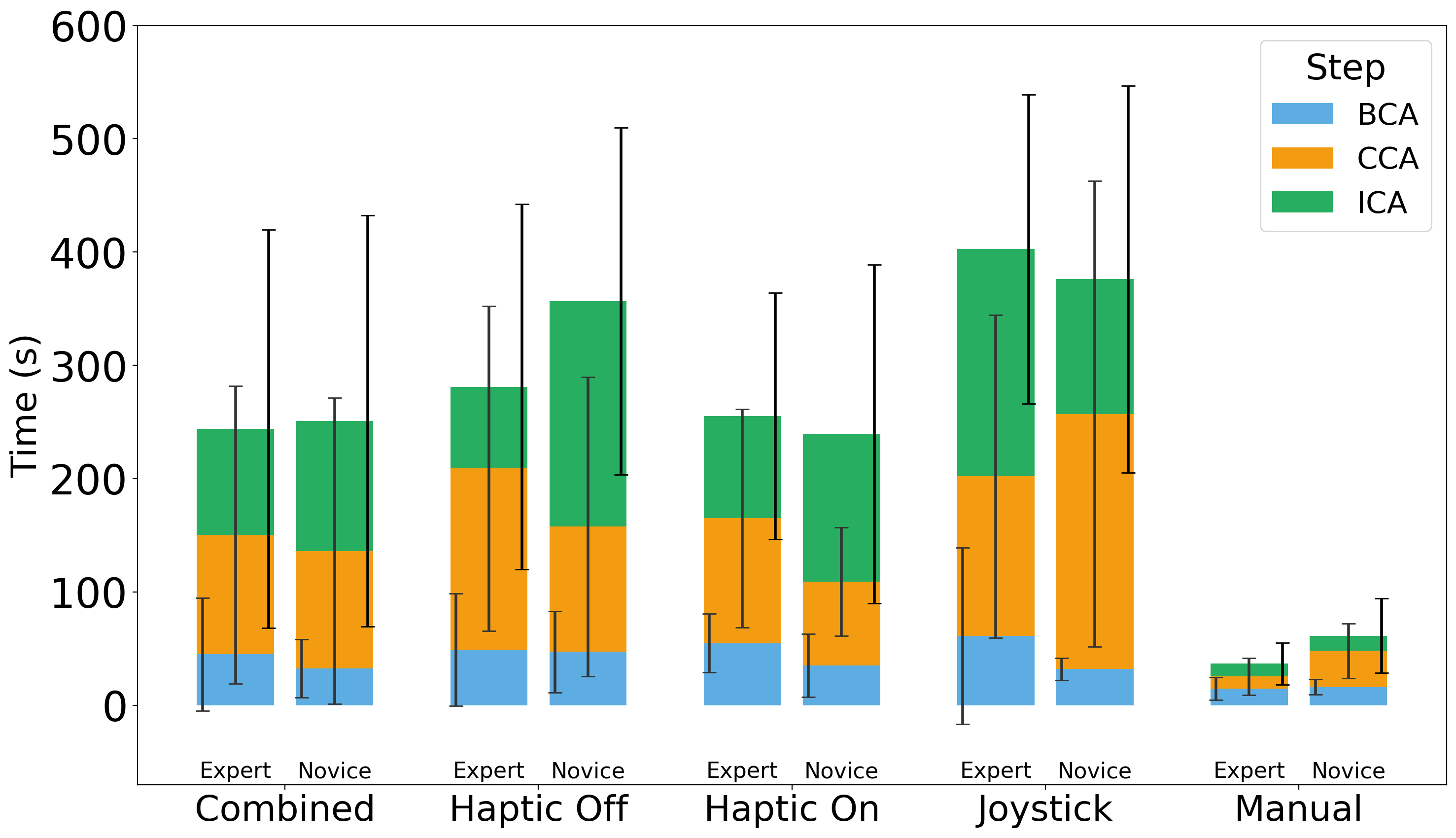}
  \caption{Novice vs. expert comparisons of checkpoint navigation time for each controller.}
  \label{fig:group_time_combined}
\end{figure}

\subsubsection{Time}

When comparing novices' navigations ($n=32$) and experts' navigations ($n=48$), combined navigation times across vessel segments (BCA, CCA, ICA) did not show significant group differences. 
Across all control methods for the BCA, novices required an average of 32.4\,s compared to 44.9\,s for experts ($t=-1.36$, $p=0.180$). 
In the CCA, novices averaged 136.2\,s while experts required 150.4\,s ($t=-0.43$, $p=0.666$). 
For the ICA, times were 250.9\,s for novices and 243.9\,s for experts, with no significant difference ($t=0.16$, $p=0.873$). 
Thus, although small numerical differences were observed in each segment, none reached statistical significance, suggesting broadly comparable performance between the two groups.

\paragraph*{Per controller comparison}
Breaking down ICA times by controller revealed strong effects of control method in both novices and experts (Fig.~\ref{fig:group_time_combined}). 
Across all participants, differences between controllers were highly significant (ANOVA: $F = 24.77$, $p < 0.001$). 
For experts alone, the effect remained strong ($F = 16.34$, $p < 0.001$), as it did for novices ($F = 8.70$, $p < 0.001$). 
In both groups, manual navigation was consistently fastest, joystick navigation slowest, and the haptic-enabled modes intermediate. 

\subsubsection{Forces: Novice vs. Expert}
\begin{figure}[htbp]
  \centering
  \includegraphics[width=0.47\textwidth]{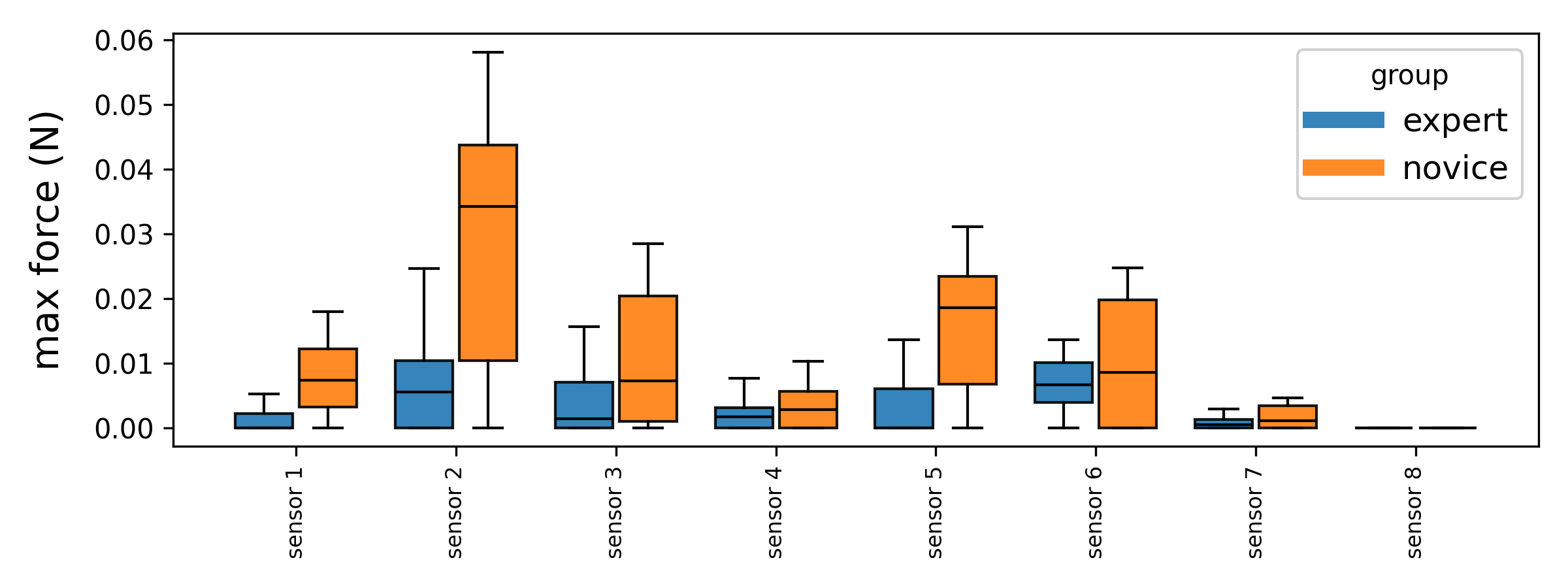}
  \caption{Comparison of maximum forces between novice and expert operators (all experiments combined).}
  \label{fig:novice_expert_overall}
\end{figure}
Maximum force data were compared between novice and expert operators on a per-sensor basis. Fig.~\ref{fig:novice_expert_overall} illustrates the distribution of forces across all controllers.

Across the eight sensors, novices consistently exerted higher forces than experts. Welch’s $t$-tests on the aggregated data indicated these sensors differences were statistically significant after Bonferroni correction ($p_{\mathrm{adj}}<0.0005$). For sensors 1, 4 and 6–8, the differences were smaller and not statistically significant. Despite novices generating higher forces, mean values for both groups remained well below the estimated puncture threshold of $0.70\pm0.29$ \cite{Siperstein2023}. This demonstrates that the robotic system maintained safe force thresholds even when operated by less experienced users.

\paragraph*{Per controller comparison}
\begin{figure}[htbp]
  \centering
  \includegraphics[width=0.47\textwidth]{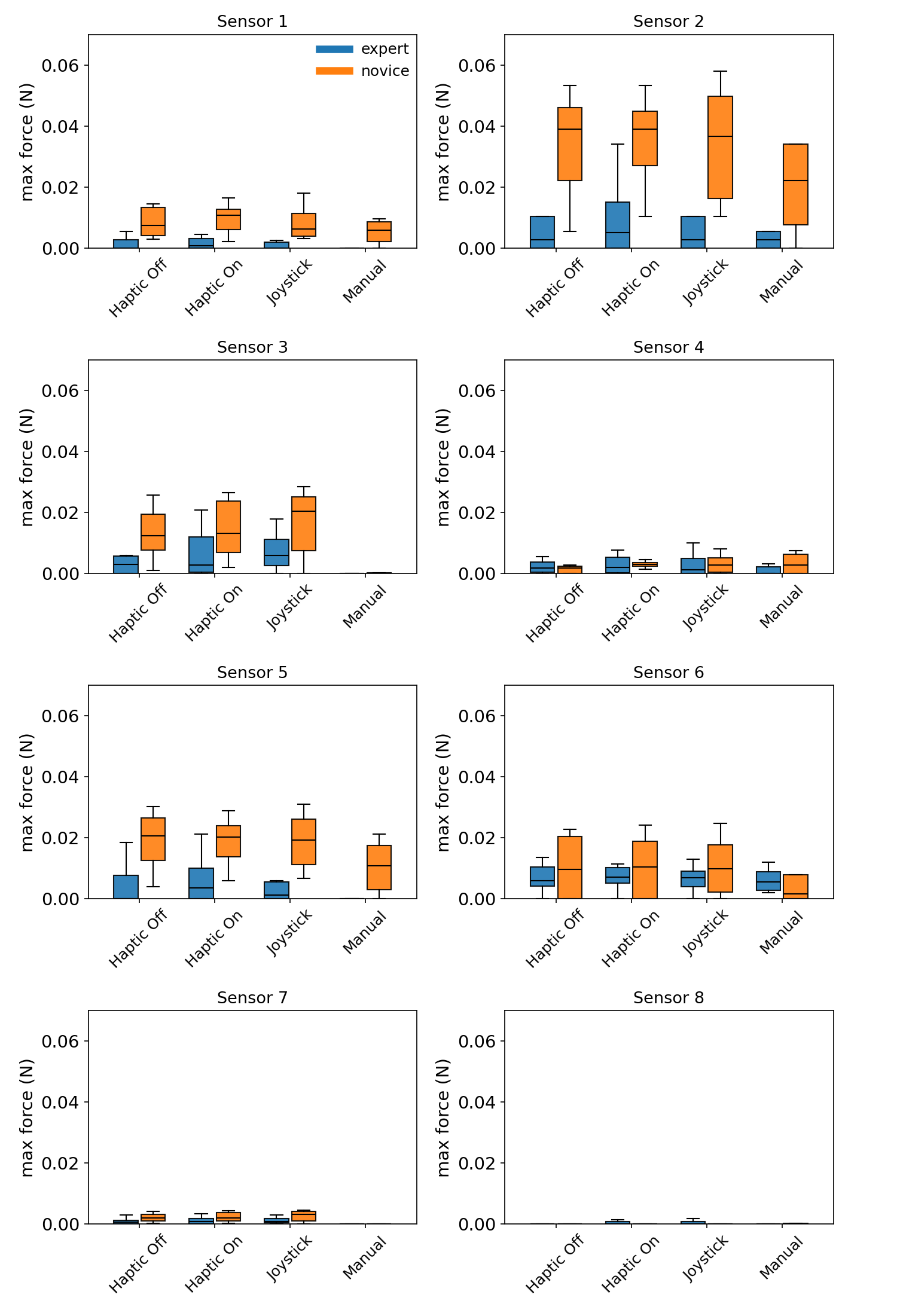}
  \caption{Per-sensor maximum force per navigation (from descending aorta to \gls{ica}) by controller and expertise group.}
  \label{fig:novice_expert_by_controller}
\end{figure}
When the analysis was stratified by control mode (Fig. \ref{fig:novice_expert_by_controller}), the same general pattern emerged. Experts exerted lower forces than novices across all controllers. Haptic on control yielded the largest novice–expert differences at sensors 1 and 5, with adjusted $p$-values of 0.012 and 0.026 respectively; these were the only controller-specific comparisons that remained significant. For haptics off, joystick and manual navigation, none of the sensor-level differences reached significance, though experts still tended to produce lower forces. Manual navigation consistently delivered the lowest absolute forces in both groups, while differences among the robotic controllers were modest.


\subsubsection{Incorrect Catheterizations: Novice vs. Expert}
\label{subsubsec:group_catheterisations}
\begin{figure}[htbp!]
  \centering
  \includegraphics[width=0.47\textwidth]{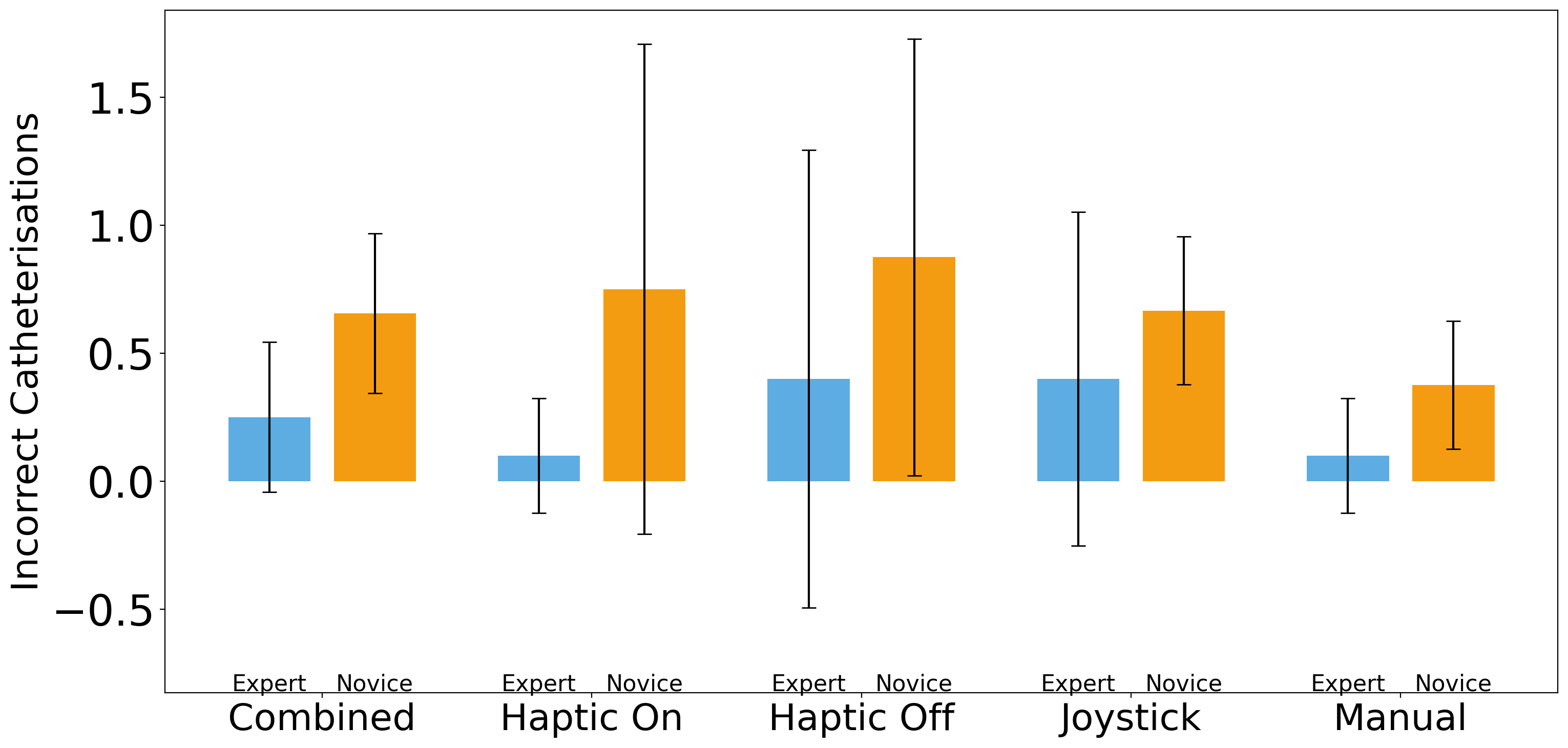}
  \caption{Novice vs. expert comparisons of incorrect catheterizations per navigation (from descending aorta to \gls{ica}).}
  \label{fig:group_catheterisations_combined}
\end{figure}
Overall, novices made significantly more incorrect catheterizations than experts, averaging 0.62 versus 0.25 errors per navigation ($p = 0.035$; Fig.~\ref{fig:group_catheterisations_combined}). This effect was consistent across controllers, indicating that operator experience plays an important role in avoiding unintended vessel entry.

\paragraph*{Per controller comparison}
When broken down by controller type, no statistically significant effect was observed (ANOVA: $F = 1.20$, $p = 0.315$). 
However, descriptive trends revealed clear differences in error rates. 
Across all participants, haptic off produced the highest mean rate of incorrect catheterizations (0.61 per navigation), followed by joystick (0.50). 
Haptic on and manual navigation showed lower error rates (0.29 and 0.22, respectively).  

Within-controller comparisons between novices and experts showed that novices consistently made more incorrect catheterizations, though differences did not reach statistical significance. For example, with the haptic on controller novices averaged 0.75 errors compared to 0.10 for experts ($t=1.33$, $p=0.269$), while in manual navigation novices averaged 0.38 versus 0.10 for experts ($t=1.72$, $p=0.135$). Similar non-significant trends were observed for haptic off and joystick. Although underpowered, these analyses suggest that operator experience may reduce errors across all controllers, with manual navigation remaining the most robust and haptic off the most error-prone.  

\subsubsection{Prolapse Events: Novice vs. Expert}
\label{subsubsec:group_prolapse}
\begin{figure}[htbp!]
  \centering
  \includegraphics[width=0.47\textwidth]{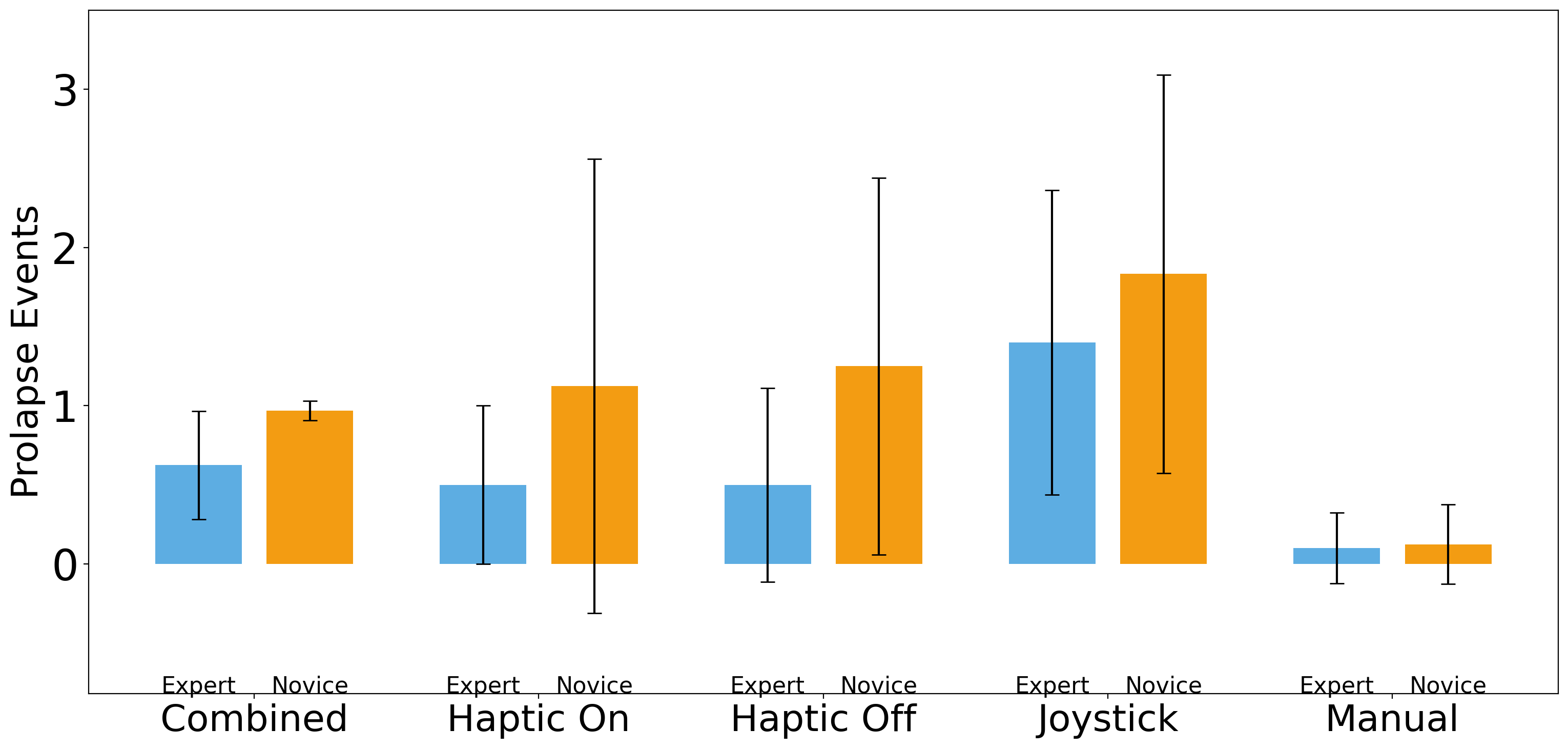}
  \caption{Novice vs. expert comparisons of prolapse events per navigation (from descending aorta to \gls{ica}).}
\end{figure}
Novices exhibited a higher mean rate of prolapse events compared to experts, averaging 0.97 versus 0.63 events per navigation (from descending aorta to \gls{ica}). Although this difference did not reach statistical significance ($t = 1.33$, $p = 0.190$), the trend suggests that operator experience may help reduce the likelihood of prolapse events.

\paragraph*{Per controller comparison}
When comparing novices and experts within each controller, novices consistently made more errors, but none of these differences reached statistical significance. For instance, novices averaged 1.25 prolapses with the haptic off controller compared to 0.50 for experts ($p = 0.312$), while in the joystick controller novices recorded 1.83 versus 1.40 for experts ($p = 0.639$). Similar non-significant differences were seen in the haptic on and manual controllers. While controller design affected prolapse frequency (Section \ref{subsec:control_verification}), these further results do not demonstrate significant differences. However, there is a pattern that operator experience may reduce prolapse frequency across all controllers. A non-significant trend further suggests that haptic feedback may reduce the novice–expert performance gap ($p = 0.133$).

\subsection{Qualitative Results}
\vspace{-0.5\baselineskip}
\paragraph{Preference for Controllers}
Overall, responses were split: four participants favored the haptic on controller, four chose haptic off, and two selected the joystick. Novices were evenly divided between haptic on and haptic off (two votes each), whereas experts were split across all three.

\paragraph{Controller Intuitiveness}
Manual navigation was rated as the most intuitive control method, with both groups giving it very high scores (overall mean $9.10 \pm 0.99$). Among the robotic interfaces, haptic on and haptic off were perceived as similarly intuitive ($6.64$ and $6.67$ respectively) and more intuitive than the joystick ($p = 0.033$). Novices generally rated all controllers as more intuitive than experts, with the ranking order remaining consistent.

\paragraph{Realism and Training Utility}
Participants rated the visual realism of the fluoroscopic images and phantom anatomy as high (means $8.27$--$9.55$), with experts assigning higher scores than novices for vascular anatomy realism ($9.00 \pm 1.00$ vs. $7.67 \pm 1.63$). There was moderate support from both groups that the platform would be a good teaching and practice tool (mean $6.55 \pm 2.98$).

\paragraph{Workload (NASA-TLX) and Haptic Effects}
No significant difference can be found between haptic on and haptic off. However, when comparing the device mimicking control with the joystick control, operators tended to rate their performance higher with the device mimicking control ($p = 0.034$)  

\paragraph{System Usability (SUS)}
The overall SUS score was applied to individual’s preferred controller and was  high ($90.91 \pm 30.50$), corresponding to a “Grade B” rating. Experts gave slightly higher scores than novices ($95.00 \pm 32.00$ vs. $85.00 \pm 28.00$), indicating greater confidence and satisfaction with the system. Combined with the workload and intuitiveness data, this suggests that the preferred controller was generally perceived as usable and effective.

\paragraph{Novice vs. Expert Comparison}
Novices consistently reported higher mental and physical demand and lower confidence in performance than experts, particularly with the joystick. Although non-significant ($p=0.56$), both groups reported slightly higher effort required with the joystick navigation, especially experts who reported an effort score of $6.80 \pm 1.80$, $6.20 \pm 1.50$ and $8.00 \pm 2.00$ for the haptic on, haptic off and joystick respectively ($p = 0.11$).

\section{Discussion}
\label{sec:discussion}

This work evaluates \textit{in vitro} comparative verification of controller interfaces for robotic interventional neuroradiology. A sensorized vascular phantom was developed to quantify device–vessel wall interactions as a safety surrogate. Robotic control strategies, including haptic feedback, were compared with conventional manual control, and performance was evaluated across operators of differing experience levels.

\vspace{-0.3\baselineskip}
\clearpage
\onecolumn
\small

\begin{longtable}{@{}p{0.48\textwidth}*{3}{>{\centering\arraybackslash}p{0.16\textwidth}}@{}}
\caption{Qualitative outcomes: Overall vs.\ Novice ($\leq$5y) vs.\ Expert ($>$5y). 
Values are mean $\pm$ SD from a 10 point Likert scale.}

\label{tab:qual_master}\\
\toprule
\thead{Item} & \thead{Overall (n = 10)} & \thead{Novice (n = 4)} & \thead{Expert (n = 6)} \\
\midrule
\endfirsthead

\toprule
\thead{Item} & \thead{Overall} & \thead{Novice} & \thead{Expert} \\
\midrule
\endhead

\midrule
\multicolumn{4}{r}{\emph{Continued on next page}}\\
\bottomrule
\endfoot

\endlastfoot

\multicolumn{4}{@{}l}{\textbf{Perceived Realism}}\\[2pt]
\makecell{The system provided a realistic fluoroscopic\\image of the vascular anatomy.} & 8.27 \(\pm\) 1.49 & 7.67 \(\pm\) 1.63 & 9.00 \(\pm\) 1.00 \\
\makecell{The system provided a realistic fluoroscopic\\ image of the catheter and guidewire.} & 9.55 \(\pm\) 0.69 & 9.33 \(\pm\) 0.82 & 9.80 \(\pm\) 0.45 \\
\makecell{The endovascular phantom anatomy was\\ realistic.} & 8.45 \(\pm\) 1.04 & 8.33 \(\pm\) 1.03 & 8.60 \(\pm\) 1.14 \\
\addlinespace[4pt]


\multicolumn{4}{@{}l}{\textbf{Controller Intuitiveness}}\\[2pt]
\hspace{1em}Manual    & 9.10 \(\pm\) 0.99 & 9.00 \(\pm\) 1.22 & 9.20 \(\pm\) 0.84 \\
\hspace{1em}Haptic On & 6.64 \(\pm\) 1.80 & 6.67 \(\pm\) 1.75 & 6.60 \(\pm\) 2.07 \\
\hspace{1em}Haptic Off & 6.67 \(\pm\) 1.50 & 7.25 \(\pm\) 2.22 & 6.20 \(\pm\) 0.45 \\
\hspace{1em}Joystick  & 5.55 \(\pm\) 2.25 & 5.83 \(\pm\) 2.71 & 5.20 \(\pm\) 1.79 \\
\addlinespace[6pt]

\multicolumn{4}{@{}l}{\textbf{Training Utility}}\\[2pt]
\makecell{Good teaching tool for interventional\\ radiology procedures.} & 6.55 \(\pm\) 2.98 & 6.83 \(\pm\) 2.79 & 6.20 \(\pm\) 3.49 \\
\makecell{Good practice / preparation tool for\\interventional radiology procedures.} & 6.55 \(\pm\) 2.98 & 6.83 \(\pm\) 2.79 & 6.20 \(\pm\) 3.49 \\
\addlinespace[4pt]

\multicolumn{4}{@{}l}{\textbf{NASA\textendash TLX (by controller)}}\\
\multicolumn{4}{@{}l}{\hspace{1em}\textit{Manual}}\\
Mental Demand      & 3.70 \(\pm\) 2.53 & 5.40 \(\pm\) 2.58 & 2.00 \(\pm\) 0.63 \\
Physical Demand    & 2.80 \(\pm\) 2.49 & 5.00 \(\pm\) 2.00 & 1.00 \(\pm\) 0.50 \\
Temporal Demand    & 3.50 \(\pm\) 2.88 & 5.50 \(\pm\) 2.90 & 2.00 \(\pm\) 0.50 \\
Performance        & 8.20 \(\pm\) 1.87 & 7.00 \(\pm\) 2.00 & 9.00 \(\pm\) 1.00 \\
Effort             & 4.00 \(\pm\) 2.40 & 5.50 \(\pm\) 2.00 & 2.50 \(\pm\) 1.00 \\
Frustration        & 2.60 \(\pm\) 2.41 & 4.00 \(\pm\) 2.00 & 1.00 \(\pm\) 0.50 \\
\addlinespace[2pt]
\multicolumn{4}{@{}l}{\hspace{1em}\textit{Haptic On}}\\
Mental Demand      & 6.64 \(\pm\) 1.69 & 6.67 \(\pm\) 1.97 & 6.60 \(\pm\) 1.14 \\
Physical Demand    & 3.09 \(\pm\) 1.58 & 3.00 \(\pm\) 1.60 & 3.10 \(\pm\) 1.50 \\
Temporal Demand    & 5.09 \(\pm\) 2.47 & 5.00 \(\pm\) 2.50 & 5.10 \(\pm\) 2.40 \\
Performance        & 5.82 \(\pm\) 2.09 & 5.50 \(\pm\) 2.00 & 6.00 \(\pm\) 2.00 \\
Effort             & 6.82 \(\pm\) 1.83 & 7.00 \(\pm\) 2.00 & 6.80 \(\pm\) 1.80 \\
Frustration        & 5.44 \(\pm\) 2.50 & 5.50 \(\pm\) 2.50 & 5.36 \(\pm\) 2.50 \\
\addlinespace[2pt]
\multicolumn{4}{@{}l}{\hspace{1em}\textit{Haptic Off}}\\
Mental Demand      & 6.45 \(\pm\) 1.56 & 6.67 \(\pm\) 1.80 & 6.20 \(\pm\) 1.17 \\
Physical Demand    & 3.27 \(\pm\) 1.90 & 3.50 \(\pm\) 2.00 & 3.00 \(\pm\) 1.50 \\
Temporal Demand    & 5.36 \(\pm\) 2.29 & 5.00 \(\pm\) 2.30 & 5.50 \(\pm\) 2.50 \\
Performance        & 6.00 \(\pm\) 1.90 & 5.50 \(\pm\) 2.00 & 6.50 \(\pm\) 1.50 \\
Effort             & 6.45 \(\pm\) 1.81 & 7.00 \(\pm\) 2.00 & 6.20 \(\pm\) 1.50 \\
Frustration        & 5.09 \(\pm\) 2.63 & 5.67 \(\pm\) 2.70 & 5.00 \(\pm\) 2.40 \\
\addlinespace[2pt]
\multicolumn{4}{@{}l}{\hspace{1em}\textit{Joystick}}\\
Mental Demand      & 6.73 \(\pm\) 1.29 & 6.67 \(\pm\) 1.37 & 6.80 \(\pm\) 1.17 \\
Physical Demand    & 2.91 \(\pm\) 2.02 & 3.00 \(\pm\) 2.00 & 2.80 \(\pm\) 2.00 \\
Temporal Demand    & 6.00 \(\pm\) 3.10 & 6.00 \(\pm\) 3.00 & 6.00 \(\pm\) 3.10 \\
Performance        & 4.91 \(\pm\) 2.43 & 3.67 \(\pm\) 1.86 & 6.40 \(\pm\) 2.30 \\
Effort             & 7.00 \(\pm\) 2.32 & 6.00 \(\pm\) 2.00 & 8.00 \(\pm\) 2.00 \\
Frustration        & 6.18 \(\pm\) 2.14 & 5.00 \(\pm\) 2.00 & 7.20 \(\pm\) 2.00 \\
\addlinespace[6pt]

\multicolumn{4}{@{}l}{\textbf{System Usability Scale (SUS)}}\\[2pt]
Preferred controller (haptic on = \textbf{H}, haptic off = \textbf{O}, joystick = \textbf{J}) & \textbf{H} = 4, \textbf{O} = 4, \textbf{J} = 2 & \textbf{H} = 2, \textbf{O} = 2, \textbf{J} = 0 & \textbf{H} = 2, \textbf{O} = 2, \textbf{J} = 2\\
SUS score, 0–-100 & 90.91 \(\pm\) 30.50 & 85.00 \(\pm\) 28.00 & 95.00 \(\pm\) 32.00 \\

\end{longtable}

\normalsize

\twocolumn

\subsection{Study Findings}

\paragraph*{\textbf{Controller comparison}} Manual navigation was the fastest way to navigate with the fewest prolapses, incorrect catheterizations, and caused the least force on the vessel wall. Among the robotic modes, haptics on using a device-mimicking controller showed a consistent trend toward shorter navigation times than using the same device with haptics off or when using joystick control, although these differences did not reach statistical significance. Joystick control was the slowest of the three robotic interfaces and, in the qualitative data, the least favored.

In terms of safety, all robotic controllers operated with wall forces far below the puncture threshold (0.70 N), and maximal forces were broadly similar across the three robotic modes. Navigation error measures reflected the same hierarchy: the joystick produced the highest rates of prolapse and, to a lesser extent, incorrect catheterizations, with haptics on and haptics off in between. While incorrect catheterizations did not differ significantly between controllers, prolapse events were significantly more frequent with the joystick than with manual control. Taken together, these results suggest that the robotic interfaces are safe within the tested paradigm. Speed and navigation error are improved with device-mimicking controllers compared to joystick controllers. Haptic feedback may confer marginal performance advantages over non-haptic robotic control in device-mimicking controllers, albeit this is not proven.

\paragraph*{\textbf{Novices versus experts}}
Navigation times were similar between novices and experts when averaged across controllers, demonstating that controller design exerted a stronger influence on speed than operator seniority. Force analyses showed that experts generally applied less force to the wall than novices regardless of controller, however, haptic on control yielded the largest novice-expert differences. Navigation error analyses suggested experts performed with less error regardless of controller type.

\paragraph*{\textbf{User preference, intuitiveness, and workload}}
Subjective data converged broadly with the quantitative results. Manual navigation was rated most intuitive. Among robotic modes, the device-mimicking controller with haptics on and off was perceived as more intuitive than the joystick. Preferences between controller type were less consistent. The divergence between workload ratings and controller preference likely reflects variation in operator confidence: although NASA-TLX scores showed no significant difference between haptic modes, participants expressed distinct preferences (four chose each), suggesting factors beyond workload influenced choice. The overall SUS score suggested good perceived usability of the preferred controller which in 80\% of cases was a device-mimicking controller.

\paragraph*{\textbf{Force validation}} As an external safety benchmark, Pavicic et\,al.\ measured the force required to puncture small arteries and reported \(0.70 \pm 0.29~\mathrm{N}\) for a 27\text{-}gauge needle~\cite{Siperstein2023}. Using this threshold, our wall\text{-}force measurements were generally $<1\%$ of the puncture force ($<0.007~N$), and even the single highest reading remained $<15\%$ ($<0.105~\mathrm{N}$), indicating a wide safety margin for all robotic modalities evaluated. Although the Pavicic study examined facial arteries rather than cerebrovascular tissue, it provides a conservative arterial puncture reference and supports the conclusion that our platform operated well within safe mechanical limits.

\subsection{Comparison to Other Studies}

Our work builds on sensorized training platforms (Fischer et al.), force-based skill metrics (Rafii-Tari et al.), and robot-mediated haptic control (Li et al.). Like Fischer et al.~\cite{Fischer2023}, we employed 3D-printed vasculature with piezoresistive sensors; however, we focused on tortuous neurovascular anatomy, employed distributed force-sensing for localization, and specifically investigated robotic controllers with and without haptic feedback.

Rafii-Tari et al.~\cite{RafiiTari2017} discriminated operator experience using force features; our distributed sensing at high-risk vascular sites similarly revealed distinct novice-expert patterns of force and error.

Li et al.~\cite{Li2022} demonstrated reduced workload and shorter times with haptics enabled. While our device-mimicking controller showed trends favoring haptic feedback, these were not statistically significant, and our qualitative analysis did not comprehensively support haptic benefits in this paradigm. 

Clinically, the CorPath GRX robotic system has emerged as a platform for neuroendovascular interventions, with recent applications in mechanical thrombectomy ~\cite{Sajja2020}; our in~vitro comparative framework complements such clinical systems by providing controlled evaluation of controller interfaces and feedback modalities. 

\subsection{Methodological Considerations and Limitations}
Sample size (n=10, $\sim$8\% of UK workforce) limits statistical power for between-group comparisons; findings require validation in larger cohorts.  

A second limitation relates to the testbed realism. Although the phantom's fluoroscopic appearance was considered realistic by users, the radiolucent lumen submerged in glycerin with external barium markers differed from clinical angiography. Static glycerin was a practical blood analog but lacks dynamic flow for clinically representative visualization.

A third limitation is that we evaluated a single anatomy and navigation pathway to the right \gls{ica}. Broader generalization requires testing multiple anatomies and pathways. Additionally, the robotic platform and controller interfaces were tested on a single in~vitro phantom design; generalization to different robotic systems (e.g., alternative master-slave configurations, different mechanical arms, or varied haptic rendering approaches) would strengthen the applicability of findings across diverse clinical robotic platforms.

\subsection{Implications for Robotic Neurointervention}
Device-mimicking control outperforms joystick interfaces in speed, safety, and user acceptance. Haptic feedback may confer benefits in complex anatomies or procedure-specific steps (e.g., clot engagement) where subtle tactile cues are informative.

\subsection{Future work}
Future studies should increase participant numbers to better resolve haptic effects; refine the haptic pipeline (higher update rates and model-based rendering); extend testing to multiple anatomies and MT-specific maneuvers where haptic cues may be most salient; and leverage granular force maps for real-time, site-specific safety overlays during navigation.

\section{Conclusion}
This work has demonstrated that a device-mimicking robotic controller interface can facilitate safe neuroendovascular navigation and is favored over joystick controllers. Although manual navigation remained faster with fewer navigation errors, the additional time required for robotic navigation (seconds) is negligible when compared to transfer times to hospitals capable of performing MT. As such, with tele-operated robotic MT (hours), the difference in procedural speed and associated navigation errors are unlikely to compromise clinical applicability given that the interface is safe and users find it acceptable and usable.

\section*{Acknowledgment}

The authors thank the Surgical \& Interventional Engineering staff Duane James, Gayathri Nantharatnam, and Marty Rajaratnam for their support in facilitating this experiment.

\section*{Funding}

This work was supported in part by the Wellcome Trust under Grant 203148/A/16/Z and in part by the Engineering and Physical Sciences Research Council Doctoral Training Partnership under Grant EP/R513064/1.

\section*{Conflict of Interest}

The authors declare no conflict of interest related to this work.

\printbibliography

\end{document}